\documentclass[runningheads]{llncs}
\usepackage{comment}
\usepackage{graphicx}
\usepackage{amsmath}
\usepackage{cite}

\begin{document}
\title{A Hybrid Approach Towards Two Stage Bengali Question Classification Utilizing Smart Data Balancing Technique}
\titlerunning{Two Stage Bengali QC Utilizing Data Balancing Technique}
%
\author{Md. Hasibur Rahman \and
Chowdhury Rafeed Rahman \and
Ruhul Amin \and
Md. Habibur Rahman Sifat \and
Afra Anika}
\authorrunning{Md. Hasibur Rahman et al.}
%
\institute{United International University, Dhaka, Bangladesh \\
\email{mrahman161260@bscse.uiu.ac.bd, }
\email{rafeed@cse.uiu.ac.bd, }
\email{ruhulamin6678@gmail.com, }
\email{msifat152028@bscse.uiu.ac.bd,  }
\email{aanika161034@bscse.uiu.ac.bd}
}
\maketitle              
\begin{abstract}
Question Classification (QC) system classifies the questions in particular classes so that Question Answering (QA) System can provide correct answers for the questions. We present a two stage QC system for Bengali. One dimensional convolutional neural network (CNN) based model has been constructed for classifying questions into coarse classes in the first stage which uses word2vec feature representation of each word. A smart data balancing technique has been implemented in this stage which is a plus for any training dependent classification model. For each coarse class classified in the first stage, a separate Stochastic Gradient Descent (SGD) based classifier has been used in order to differentiate among the finer classes within that coarse class in stage two. TF-IDF representation of each word has been used as feature for each SGD classifier separately. Experiments show the effectiveness of this two stage classification method for Bengali question classification.   

\end{abstract}
\section{Introduction}
Question Classification (QC) system categorizes questions asked in natural language into different classes. With the increasing significance of information, people are using search based tools more robustly than ever before. These search based and knowledge based tools often retrieve appropriate information using some sort of QA system. The precondition of a sound QA system is a sound QC system. Different algorithms have been proposed in \cite{banerjee2012bengali,banerjee2013empirical,banerjee2013ensemble}, \cite{islam2016word} and \cite{nirob2017question} for Bengali question classification task.

We propose a two stage approach for Bengali QC. Our approach shows superior performance compared to state-of-the-art Bengali question classifiers. We also propose a natural language related sample augmentation technique in order to remove class imbalance by generating theoretical samples. Such augmentation technique can be of use when training deep learning based data hungry classifiers related to natural language.

In the first stage, we classify the given question into one of the six coarse classes of our dataset. In the next stage, we classify the question sample into one of the finer classes existing within the coarse class obtained from stage one. We have class imbalance among stage one coarse classes which we resolve by constructing samples using SMOTe (Synthetic Minority Oversampling Technique). We implement this technique on a special vector representation of our question samples in order to gain theoretical representative samples for the minority classes. Our Experimental results show that our approach is successful in creating representative theoretical samples from existing minority class samples. Such a balanced dataset has helped our 1D CNN based model of stage one to gain excellent results. 1D CNN works with the help of word2vec representation of each word. For each coarse class, we have a separate SGD classifier to classify the question sample into one of the finer classes within that coarse class. We have used TF-IDF representation of question words as source of features for our SGD classifiers. We keep stop words of question samples while classifying. In \cite{anika2019comparison}, the authors showed superior performance of different classifiers when stop words were not removed.    

\section{Literature Review}
\subsection{Existing Question Answering Systems} 
The oldest QA system named 'BASEBALL' \cite{green1961baseball} was developed in 1961. This system answers questions only related to baseball games. Another old QA system is 'LUNAR' \cite{woods1972lunar}. It was developed in 1972 and could answer questions about soil samples. Different QA Systems have been developed in different languages such as Arabic QA system named 'AQAS' \cite{mohammed1993knowledge}, Arabic factoid question answering system \cite{fareed2014syntactic}, Chinese QA system \cite{yu2005modified}, Hindi QA system for E-learning Documents \cite{kumar2005hindi} and Hindi - English QA system \cite{sekine2003hindi}. Various analysis procedures for QA system exist such as morphological analysis \cite{hovy2000question}, syntactical analysis \cite{zheng2002answerbus}, semantic analysis \cite{wong2007practical} and expected answer Type analysis \cite{benamara2004cooperative}. Other popular QA systems are Apple Siri, Amazon Alexa and IBM Watson. 

\subsection{Research Works on Question Classification System}
 Question classification (QC) can be performed using two approaches such as rule-based approach and machine learning based approach \cite{banerjee2012bengali}. Grammar coded rules are used to classify the questions to appropriate answer type in rule-based approach \cite{prager1999use}, \cite{voorhees1999trec}. Harish Tayyar Madabushi and Mark Lee also proposed  a rule-based approach \cite{madabushi2016high} for QC. At first, based on a question structure they extracted relevant words. Then they classified questions based on rules that associated these words to concepts.
Different types of classifiers have been used to categorize each question into a suitable answer type in machine learning based approach. Some of the examples are - Support Vector Machine (SVM) \cite{nirob2017question}, \cite{zhang2003question}, Support Vector Machines and Maximum Entropy Model \cite{huang2008question}, Naive Bayes (NB), Kernel Naive Bayes (KNB), Decision Tree (DT) and Rule Induction (RI) \cite{banerjee2012bengali}. In \cite{aouichat2018arabic}, the authors proposed an approach for QC that combined SVM model and  CNN model. In \cite{day2007question}, the authors proposed an integrated genetic algorithm (GA) and machine learning (ML) approach for question classification in English-Chinese cross-language question answering. The authors of \cite{mohasseb2018question} analyzed and marked out different patterns of questions based on their grammatical structure and used machine learning algorithms to classify them. In \cite{li2017semi}, the authors proposed a kind of semi-supervised QC method that was based on ensemble learning. Information gain and sequential pattern for feature extraction to classify English questions were used in \cite{liu2018feature}. QC systems have been developed for different languages such as Chinese language \cite{zheng2005chinese}, \cite{yu2005modified}, \cite{xu2006syntactic}, Spanish language \cite{cumbreras2006bruja}, Japanese language \cite{dridan2007classify}, Arabic Language \cite{hasan2018combined} and so on.

\subsection{Question Classification Systems in Bengali Language}
The authors extended their work done in \cite{banerjee2012bengali} from single layer taxonomy to two-layer taxonomy using 9 coarse-grained classes and 69 fine-grained classes for Bengali question classification \cite{banerjee2013ensemble}. In \cite{islam2016word}, the authors used 6 coarse classes and 50 finer classes following the method proposed in \cite{li2005semantic}. Lexical features and syntactical features were used to classify questions into appropriate classes \cite{nirob2017question}. In \cite{anika2019comparison}, the authors provided a comparison of machine learning-based methods based on performance and computational complexity. They used 7 different classifiers to conduct comparison where Stochastic Gradient Descent (SGD) performed the best.

\section{Our Dataset}
We have used the same dataset as \cite{islam2016word}. There are 3333 "wh" type of questions in the dataset. The two types of classes making up this dataset are - coarse class and finer class (within each coarse class). This aspect has been shown in Table \ref{Table:dataset}. The maximum word number of a question is 21.

\begin{table}[h]
\begin{flushleft}
\label{my-label}
\caption{Coarse and Fine Grained Question Categories}
\begin{tabular}{|p{4cm}|p{8cm}|}
\hline
\textbf{Coarse Class} & \textbf{Finer Class}\\
\hline
ENTITY (482) & SUBSTANCE (10), SYMBOL (11), CURRENCY (24), TERM (10), WORD (10), LANGUAGE (30), COLOR (10), RELIGION (15), SPORT (10), BODY (10), FOOD (11), TECHNIQUE (10), PRODUCT (10), DISEASE (10), OTHER (22), LETTER (10), VEHICLE (11), PLANT (12), CREATIVE (216), INSTRUMENT (10), ANIMAL (10), EVENT (10)\\
\hline
NUMERIC (889) & COUNT (213), DISTANCE(13),CODE(10), TEMPERATURE (13), WEIGHT (20), MONEY (10), PERCENT (27), PERIOD (33), OTHER (34), DATE (452), SPEED (10), SIZE (54)\\
\hline
HUMAN (651) & INDIVIDUAL (610), GROUP (18), DESCRIPTION (13), TITLE (10) \\
\hline
LOCATION (611) & MOUNTAIN (23), COUNTRY (105), STATE (88), OTHER (121), CITY (274) \\
\hline
DESCRIPTION (198) & DEFINITION (141), REASON (26), MANNER (12), DESCRIPTION (19) \\
\hline
ABBREVIATION (502) & ABBREVIATION (489), EXPRESSION (13) \\
\hline
\end{tabular}
\end{flushleft}
\label{Table:dataset}
\end{table}

\section{Proposed Methodology}

\subsection{Method Overview}
\begin{figure}[h]
    \centering
    \includegraphics[width=\textwidth]{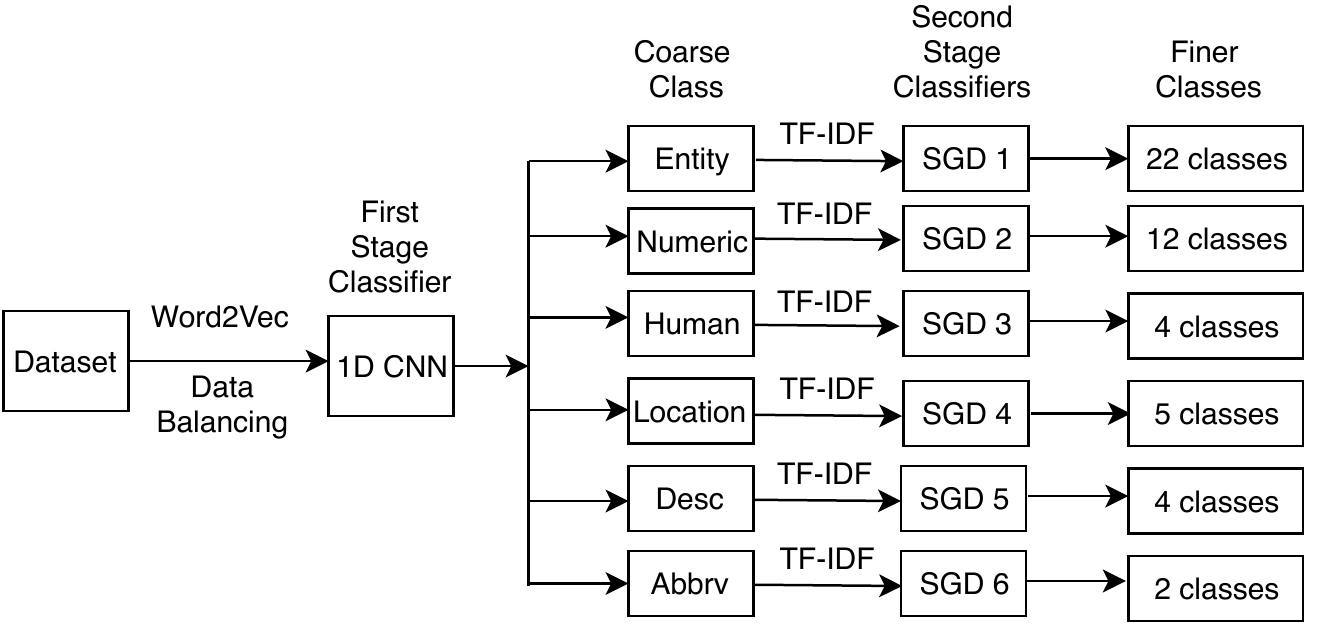}
    \caption{System High Level Overview}
\label{fig:overview}
\end{figure}

Figure \ref{fig:overview} shows the complete high level overview of our proposed technique for Bengali question classification (QC). At first, we construct word2vec representation for all words of our question corpus. Using these vector representations, we perform class balance on our six coarse class QC data. We use 1D CNN based model in order to classify a question into one of the six coarse classes in the first stage using word2vec representation as feature. In the next stage, we use a separate SGD classifier for each separate coarse class in order to classify the question into one of the finer classes residing within that coarse class. As source of feature, we use TF-IDF (Term Frequency Inverse Data Frequency) of the words residing in the question samples of each coarse class.   

\begin{figure}[h]
    \centering
    \includegraphics[width=\textwidth]{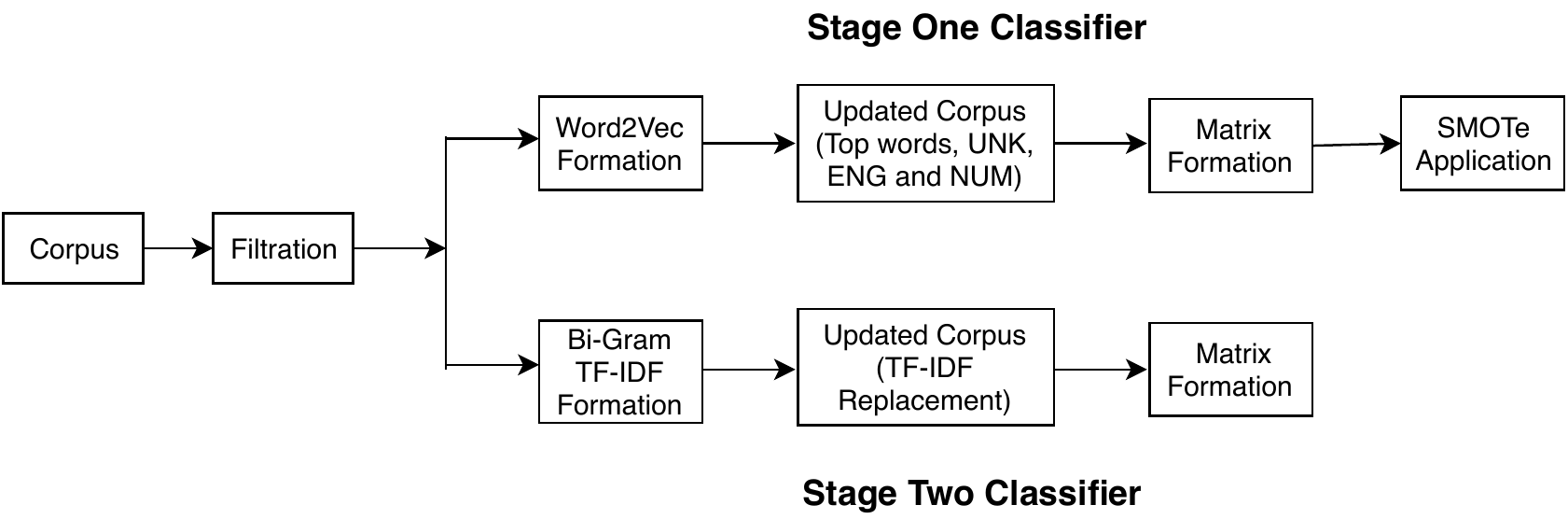}
    \caption{Data Preprocessing Overview}
\label{fig:Data}
\end{figure}

Figure \ref{fig:Data} shows the data processing steps of our system. In the \textbf{filtration} step, we remove all the punctuation [ ex: ',', '.' , '?' \ldots] from the dataset. We now take two different routes of data preprocessing for our stage one and stage two classifier as shown in the figure. These steps are described in Subsection \ref{Coarse_data} and  \ref{Finer_data}. 

\subsection{Coarse Class Classification}\label{Coarse_data}
\subsubsection{Data Preprocessing:} 
We need appropriate numeric representation of each word in order to train and test any deep learning based model. We use \textbf{word2vec} representation of each word as feature for the stage one classifier. For constructing word2vec of a particular word, a neural network hidden layer is trained such that the input is one hot vector of that word and target output is the probability distribution of all words being neighbour of our word of interest. The goal is to have similar vector representation for words of similar meaning.

We learn unique word2vec representation of only those words which appear at least 15 times in our corpus. We call these words our \textbf{top words}. There are 163 such words in our corpus. The learning process of word2vec would fail if we took non-frequent words as well. If a word represents numeric value, we replace that word with special keyword NUM. We also replace English words with ENG. Apart from these two kinds of words, words appearing less than 15 times are replaced with UNK keyword. Finally, we form word2vec of size 100 for each of the 163 top words, UNK, NUM and ENG. We store up the vectors in a new updated corpus. Each question sample is now of dimension $21\times100$ as vector size of each word is 100, and we pad each question sample such that all samples have the same length of 21 (the highest length sample has 21 words).

\subsubsection{Data Balancing:}
We apply SMOTe (Synthetic Minority Oversampling Technique) on our dataset consisting of six coarse classes in order to gain class balance. We need one dimensional samples as SMOTe is a distance based method. We flatten each of our question sample of dimension $21\times100$ turning them into 2100 size one dimensional vector. In SMOTe, we first take samples of the feature space for each target class and its nearest neighbors, and then generate new instance by combining those features. Thus we oversample our minority classes generating representative theoretical samples each containing 2100 features. We then reshape each sample to previous two dimension of $21\times100$. Our experiments prove the effectiveness of this smart theoretical sample generation process using SMOTe utilizing word2vec features. It is to note that we apply SMOTe only on training data. All of our validation samples come from the actual dataset. Such oversampling helps our data hungry convolutional neural network based model to learn class discriminating features. 

\subsubsection{CNN Model:}
\begin{figure}[h]
\centering
\includegraphics[width=0.3\textwidth]{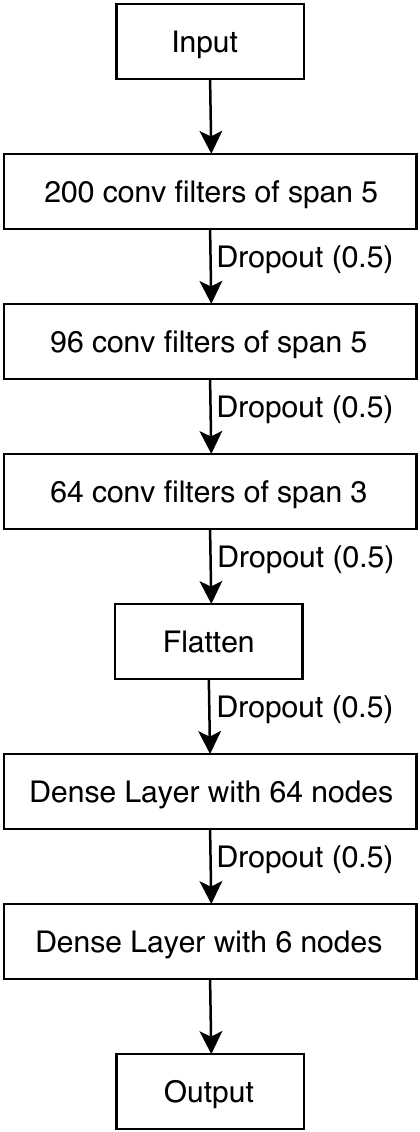}
\caption{Architecture of 1D Convolutional Neural Network}
\label{fig:1D_CNN}
\end{figure}

Figure \ref{fig:1D_CNN} shows the architecture of our 1D CNN. This is a typical 1D CNN architecture consisting of some convolution layers as the first few layers and dense layers coming at later stages. We do not use any pooling layer as we have found out a decrease of validation accuracy while using such layers. It is probably because of the loss of some useful local features while pooling. We have used dropout layers after every convolution and dense layers in order to reduce overfitting. Except for the last layer, we use \textbf{relu} activation function in each layer. In the final output layer, we use \textbf{softmax} activation function. We use \textbf{Adam} optimizer for parameter update and \textbf{Categorical Crossentropy} loss function for performing multi-class classification. CNN can extract features from local input patches allowing for data efficiency and representation modularity. These same properties make them highly significant to sequence processing.  

\subsection{Finer Class Classification}\label{Finer_data}
\subsubsection{Data Preprocessing:}
We use TF-IDF representation for numeric representation of each word in order to train and test our finer class classifiers. TF-IDF indicates Term Frequency - Inverse Data Frequency. We use bi-gram for TF-IDF construction. The goal of bi-gram TF-IDF is to assign higher weights to the word couples that are more significant for our classification process. Generally, the word couples that appear many times in one class of data and have very low frequency in other classes are the most significant. Word couples that appear frequently in samples of all classes are generally insignificant. TF-IDF is used as feature with classifiers that have low number of parameters capable of learning even from small number of training samples.

Each of the six coarse classes of our dataset are divided into finer classes. For example, \textbf{Entity} coarse class is divided into 22 finer classes. We implement TF-IDF separately for the question samples of each separate coarse class as we have separate SGD model for each separate coarse class. The words which are not frequent among all class samples are the ones that actually help in distinguishing between the classes and should carry more weight. As a result, TF-IDF ensures less weight for stop words and more weight for special keywords. We calculate TF-IDF score for each unique bi-gram of a coarse class. Then we construct a one dimensional score vector for each question sample of that coarse class. Suppose, in a particular coarse class, there are total 500 question samples and 2000 unique bi-grams. Now, each sample will be of dimension 2000 for that particular coarse class. It is because we replace each bi-gram by its TF-IDF score according to the presence or absence of that bi-gram in that sample. We do not use any data balancing technique for finer classes because of two reasons. The first reason is that stage one classifier shortlists the possible finer classes by allowing us to look into the appropriate coarse class. The second reason is that we have separate SGD classifier (can learn using small number of training samples) working on the finer classes of each separate coarse class. This allows each SGD model to specialize on the finer classes within its relevant coarse class. 

\subsubsection{SGD Classifier}
SGD (Stochastic Gradient Descent) is an iterative algorithm which is used for optimizing a particular objective function. It optimizes an unbiased function with suitable smoothing properties. For each iteration, a set of instances are chosen randomly for parameter update instead of choosing all instances in a dataset all at once. We use \textbf{Huber} loss function instead of mean squared error as it is less sensitive to anomalous data points. To reduce over-fitting, we use \textbf{L2 regularization}.

\section{Results and Discussion}
We have used 10 fold cross validation for evaluating our proposed methods, as it prevents the rise or fall of validation accuracy by chance. For performance evaluation, we use \textbf{precision}, \textbf{recall} and \textbf{f1 score}. We calculate these measures as follows: 
\noindent 
\begin{equation}
Precision = \dfrac{True Positive}{True Positive + False Positive} 
\end{equation}
\begin{equation}
Recall = \dfrac{True Positive}{True Positive + False Negative} 
\end{equation}
\begin{equation}
f1-Score = 2 * \dfrac{Precision * Recall}{Precision + Recall} 
\end{equation}

It is to note that we have not eliminated stop words. In \cite{anika2019comparison}, all the machine learning based algorithms performed better when stop words were not eliminated. 
Table \ref{Table:coarse_result} shows the average result of precision, recall and f1-score after 10 fold cross-validation for coarse class classification. The validation accuracy for coarse class classification in our case is close to 95\% which is significantly higher compared to validation accuracy of 89\% obtained from the application of SGD. 1D CNN successfully learns discriminating features for coarse class classification with the help of data balancing technique. 

\begin{table}[h]
\centering
\label{my-label}
\caption{Experiment Results of Coarse Class Classification}
\begin{tabular}{|p{2cm}|c|c|}
\hline
Precision   & 0.9310               \\ \hline
Recall      & 0.9344               \\ \hline
F1 Score    & 0.9325               \\ \hline
\end{tabular}
\label{Table:coarse_result}
\end{table}

Table \ref{Table:finer_result} shows the precision, recall and f1 score of all six SGD based models and the average of those scores. It is to note that Model 1 of the table is the SGD model used for classifying the finer classes within coarse class one (Entity coarse class). Similar implications are applicable for the other five models. Our method shows superior performance when it comes to finer class classification compared to the finer class classification results provided in \cite{anika2019comparison} (Results provided in Table \ref{Table:previous_finer_result}). This has been possible, because each of our six SGD models has the advantage of specializing on the finer classes of only one coarse class. 

\begin{table}[h]
\centering
\label{my-label}
\caption{Experiment Results of Finer Class Classification}
\begin{tabular}{|l|c|c|c|c|c|c|c|}
\hline
          & Model 1 & Model 2 & Model 3 &  Model 4 & Model 5 & Model 6 & Average \\ \hline
Precision & 0.9198    &  0.7693 &  0.8033   & 0.9035 &0.9513  &0.9282   & 0.8792        \\ \hline
Recall    & 0.9404    &  0.7586 &  0.8371   & 0.9035 &0.9641  &0.9048   & 0.8847      \\ \hline
F1 Score  & 0.9297    &  0.7404 &  0.8091   & 0.8964 &0.9565  &0.9018   & 0.8723        \\ \hline
\end{tabular}
\label{Table:finer_result}
\end{table}

In natural language based classification tasks where there are main classes and sub-classes within each main class, such two stage approach can be a good way of boosting performance. Our proposed data balancing technique can be used with any imbalanced natural language based classification dataset where data hungry deep learning based models are to be implemented. It is to note that deep learning based 1D CNN model has given poor accuracy while trained and validated on the finer classes, as number of samples of each of these classes is insufficient to train data hungry deep learning based model. In such cases, models such as stochastic gradient descent which have low number of parameters to learn should be used.  

\begin{table}[h]
\centering
\label{my-label}
\caption{Finer Class Classification Performance of State-of-the-art Approaches}
\begin{tabular}{|c|p{1.3cm}|p{1.3cm}|p{1.3cm}|p{1.3cm}|p{1.3cm}|p{1.3cm}|p{1.3cm}|}
\hline
& MLP   & SVM   & NBC   & SGD   & GBC   & KNN   & RF    \\ \hline
Accuracy & 0.83  & 0.801 & 0.789 & 0.832 & 0.792 & 0.781 & 0.816 \\ 
\hline
F1 Score & 0.810 & 0.765 & 0.759 & 0.808 & 0.775 & 0.755 & 0.783 \\ 
\hline
\end{tabular}
\label{Table:previous_finer_result}
\end{table}

\section{Conclusion}
We have introduced a two stage approach for Bengali question classification - a deep learning based approach in the first stage and a gradient descent based approach in the second stage. We have also introduced a way of creating new representative theoretical samples for each coarse class which assists in maintaining class balance in training set. We have shown the effectiveness of our approach through experiments. Researchers working on building Bengali question answering system can follow this work as part of their question classification module. Our finer class classifiers are expected to show better performance provided that more training data per finer class is collected. We leave this as part of future work.      



%

%
\bibliographystyle{splncs04}
\bibliography{main}

\end{document}